\documentclass{article}
\usepackage{spconf,amsmath,graphicx,amssymb}
\usepackage{multirow}
\usepackage{bm}
\usepackage[table,xcdraw]{xcolor}
\usepackage{url} 
\usepackage{mathrsfs}
\usepackage{booktabs}
\usepackage{hyperref}

\title{Zero-Shot Co-salient object detection Framework}
\name{Haoke Xiao$^1$ \quad Lv Tang$^{2*}$ \quad Bo Li$^2$ \quad
Zhiming Luo$^1$ \quad Shaozi Li$^{1*}$
\thanks{*Corresponding authors: Lv Tang and Shaozi Li. Email: hk.xiao.me@gmail.com, lvtang@vivo.com, libra@vivo.com, zhiming.luo@xmu.edu.cn and szlig@xmu.edu.cn. This work is supported by the National Natural Science Foundation of China (No.~62276221, No.~62376232). The code can be found
\href{https://github.com/hkxiao/zs-cosod}{\color{red}here}.
\\
}}
\address{$^1$ Institute of Artificial Intelligence, Xiamen University, Xiamen, China \\ $^2$ vivo Mobile Communication Co., Ltd, Shanghai, China}

\begin{document}
\maketitle

\begin{abstract}
Co-salient Object Detection (CoSOD) endeavors to replicate the human visual system's capacity to recognize common and salient objects within a collection of images. Despite recent advancements in deep learning models, these models still rely on training with well-annotated CoSOD datasets. The exploration of training-free zero-shot CoSOD frameworks has been limited. In this paper, taking inspiration from the zero-shot transfer capabilities of foundational computer vision models, we introduce the first zero-shot CoSOD framework that harnesses these models without any training process. To achieve this, we introduce two novel components in our proposed framework: the group prompt generation (GPG) module and the co-saliency map generation (CMP) module. We evaluate the framework's performance on widely-used datasets and observe impressive results. Our approach surpasses existing unsupervised methods and even outperforms fully supervised methods developed before 2020, while remaining competitive with some fully supervised methods developed before 2022. 
\end{abstract}


\section{Introduction}
Co-salient object detection (CoSOD) is a task that seeks to replicate the human visual system's ability to identify common and salient objects from a set of related images. One of the unique challenges in CoSOD is that co-salient objects belong to the same semantic category, but their specific category attributes remain unknown. These distinctive characteristics have made CoSOD an emerging and demanding task that has gained rapid traction in recent years~\cite{10008072,9358006,DBLP:journals/tist/ZhangFHBL18}.

\begin{figure}
\centering
\includegraphics[width=1.0\linewidth]{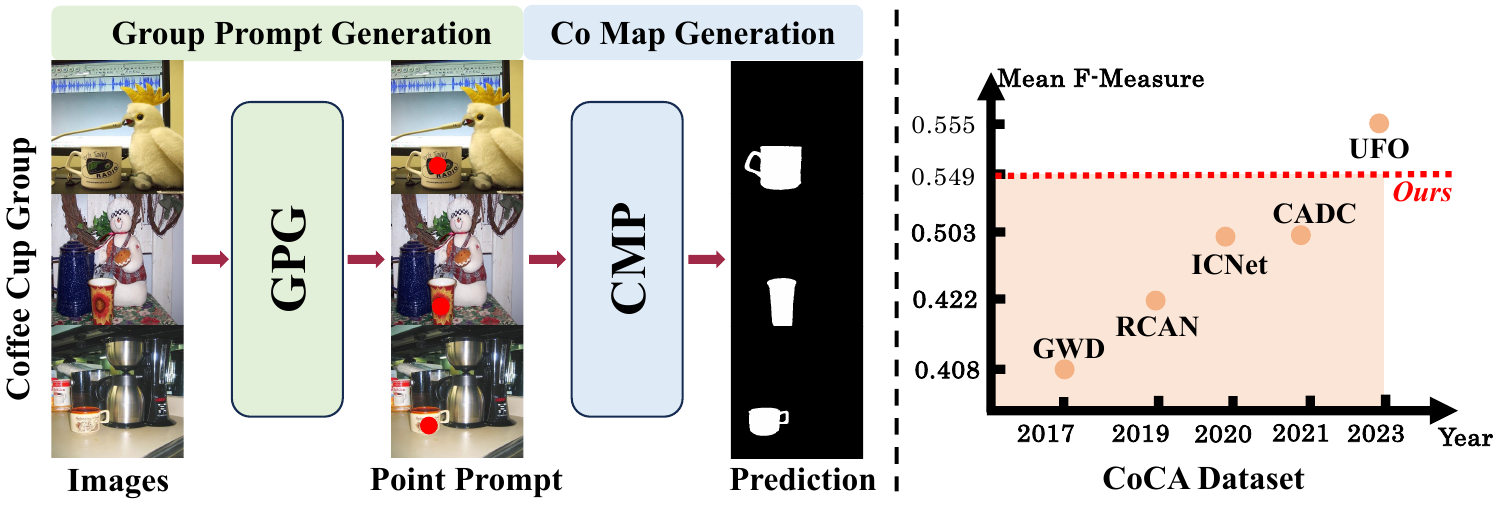}
\caption{Left: The architecture of our proposed zero-shot CoSOD framework. Right: The performance of our proposed zero-shot CoSOD framework. GWD~\cite{DBLP:conf/ijcai/WeiZBLW17}, RCAN~\cite{DBLP:conf/ijcai/0061STSS19}, ICNet~\cite{DBLP:conf/nips/Jin0CZG20}, CADC~\cite{DBLP:conf/iccv/ZhangHL021} and UFO~\cite{su2023unified} are five typical methods.}
\vspace{-0.4cm}
\label{arch_introd}
\end{figure}

Detecting co-saliency fundamentally hinges on accurately modeling the inter relation within an image group. To tackle this challenge, various meticulously designed architectures have been proposed, including RNN-based methods~\cite{DBLP:conf/ijcai/0061STSS19,DBLP:journals/tip/0115TKSD22}, CNN-based methods~\cite{DBLP:conf/nips/ZhangCHLZ20,DBLP:conf/nips/Jin0CZG20,DBLP:journals/tcsv/TangLKSD22}, and Transformer-based methods~\cite{su2023unified,DBLP:conf/cvpr/LiHZLKCAK23}. While these methods have achieved impressive performance, they often rely on small-scale datasets and require the integration of complex network modules. It's important to highlight that previous studies~\cite{10008072,DBLP:journals/tip/0115TKSD22} show that changing the training data while keeping the same network architecture, or modifying the backbone network while using the same training data can significantly impact network performance. This suggests that improving network performance might be attainable by employing a more robust backbone network or higher-quality training datasets. These considerations prompt us to reevaluate whether addressing the CoSOD task necessitates the design of intricate and diverse modules, or if an alternative approach should be explored.

Recently, foundational computer vision (CV) models, such as SAM~\cite{DBLP:journals/corr/abs-2304-02643} and DINO~\cite{DBLP:journals/corr/abs-2304-07193}, have emerged. These models, once trained, can be seamlessly applied to various downstream tasks in a zero-shot manner, eliminating the need for dataset-specific fine-tuning. This prompts us to explore whether these CV foundational models can be harnessed for CoSOD. However, existing CV foundational models, like SAM, are tailored for single-image tasks and lack the capacity to discern inter-saliency relations within an image group. Moreover, manually supplying SAM with inter-saliency or group prompts, which aid in co-saliency map generation, is impractical due to the nature of the CoSOD task.

\begin{figure*}[!htbp]
    \centering
    \includegraphics[width=0.83\linewidth]{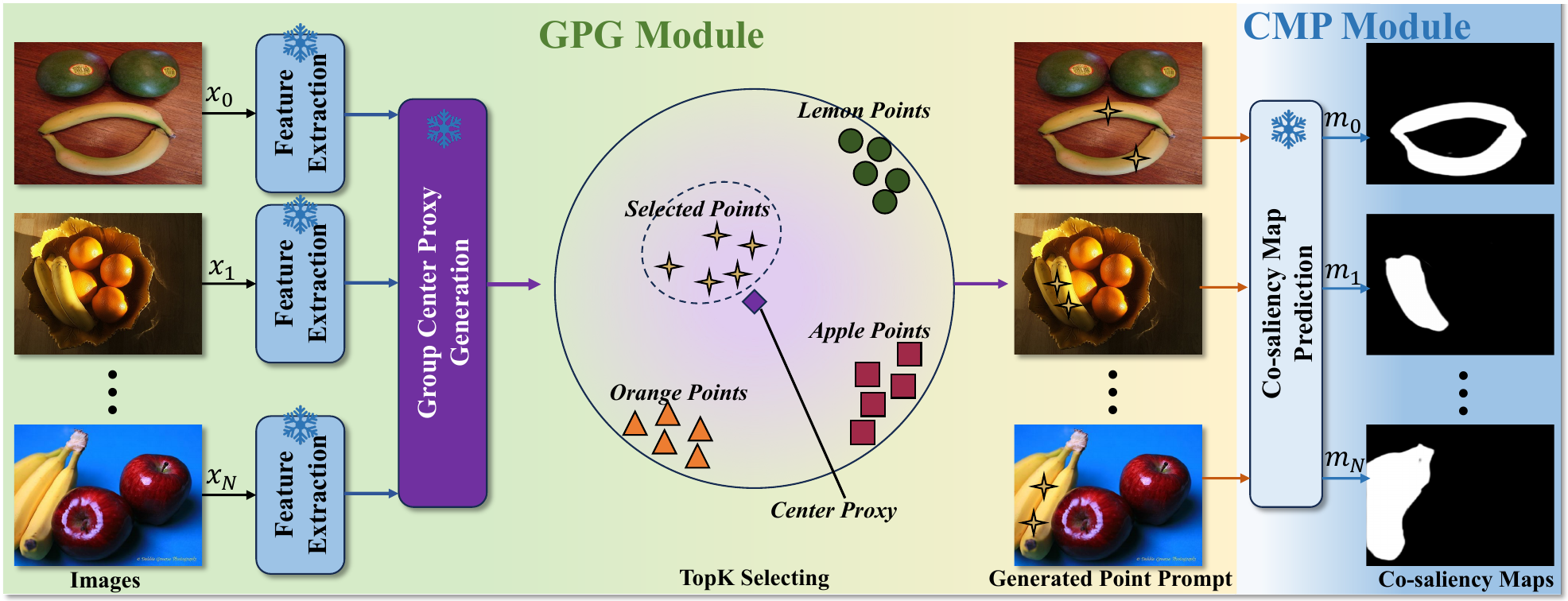}
    \caption{The architecture of our proposed zero-shot CoSOD framework. Feature extraction is accomplished by utilizing \textbf{DINO} and \textbf{SD} to extract both high-level and low-level information. The CMP module employs \textbf{SAM} to generate the co-saliency maps. Importantly, all parameters in the network remain frozen, eliminating the need for additional training.}
    \label{framework}
    \vspace{-0.5cm}
\end{figure*}

To tackle the aforementioned challenges, we present an innovative zero-shot CoSOD framework that leverages foundational computer vision models. As depicted in Fig. \ref{arch_introd}, our framework consists of two main components: group prompt generation (GPG) and co-saliency map generation (CMP). In the GPG module, we initially extract high-level semantic information from each image using the CV foundational model. We also explore the supplementation of low-level spatial details using Stable Diffusion (SD)~\cite{DBLP:conf/cvpr/RombachBLEO22}, which may not be captured by the foundational model. Subsequently, we combine these pieces of information to generate group prompts. These prompts, created by the GPG module, serve as input for the CMP module. As depicted in Fig. \ref{arch_introd}, our network surpasses methods developed before 2021. Our key contributions are:

\begin{itemize}
\item We take the pioneering step of introducing a zero-shot CoSOD framework, potentially inspiring researchers to address the CoSOD from a fresh perspective.

\item To address the limitations of existing CV foundational model when applied to CoSOD task, we further design the GPG and CMP modules.

\item We validate our zero-shot CoSOD framework on three widely used datasets (CoCA~\cite{DBLP:conf/eccv/ZhangJXC20}, CoSOD3k~\cite{9358006} and Cosal2015~\cite{DBLP:journals/ijcv/ZhangHLWL16}), and the performance shows the effectiveness of our proposed zero-shot framework.

\end{itemize}

\section{METHOD}
Our emphasis lies in providing an in-depth description of the key components within GPG. GPG encompasses feature extraction, group center proxy generation, and TopK selection.

\subsection{Feature Extraction}
\noindent \textbf{High-level Feature Extraction.}
Existing works~\cite{DBLP:conf/iccv/TruongDYG21,DBLP:journals/corr/abs-2304-07193} demonstrate that self-supervised ViT features, as exemplified by DINO, contain explicit information for semantic segmentation and excel as KNN classifiers. In essence, DINO is adept at accurately extracting the semantic content of each image, a vital aspect in discerning the group features within an image set. Herein, we choose the 11th layer feature $\mathcal{F}_{DINO}$  to represent the semantic information of each image. 

\begin{figure}[!t]
    \centering
    \includegraphics[width=0.85\linewidth]{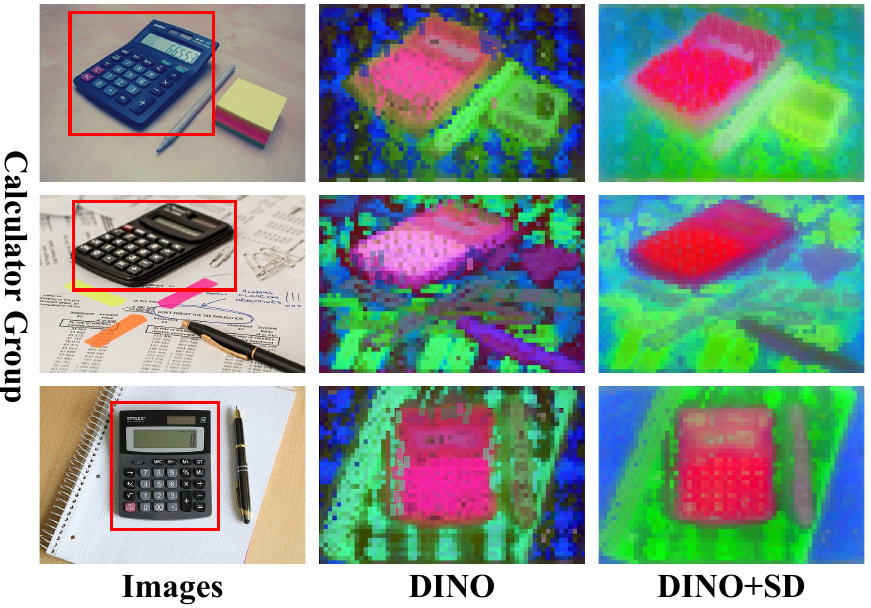}
    \caption{The generated group features.}
    \label{dinovis}
    \vspace{-0.5cm}
\end{figure}

\noindent \textbf{Low-level Feature Extraction.}
While DINO excels in providing substantial high-level semantics, it lacks in delivering nuanced low-level spatial information. In the second column of Fig. \ref{dinovis}, the group feature generated solely through DINO would lack low-level detailed information. As emphasized in previous studies~\cite{DBLP:conf/cvpr/ZhangLSLC020,DBLP:journals/tip/0115TKSD22}, both low-level and high-level features are pivotal for modeling group features. However, it's noteworthy that there is a research gap regarding the supplementation of low-level spatial information to features extracted by DINO in a zero-shot manner. 

In our proposed network, the inclusion of a pre-trained model that specializes in low-level spatial information becomes crucial, particularly in scenarios lacking strong texture cues. Such a model can effectively complement the low-level spatial information extracted by DINO. Notably, SD~\cite{DBLP:conf/cvpr/RombachBLEO22} has recently showcased its exceptional ability to generate high-quality images. This underscores its potential for robustly representing images, encompassing both content and spatial information. Consequently, our primary objective is to explore whether SD features can enhance the establishment of inter-relationships when combined with DINO. 

The architecture of SD comprises three key components: an encoder $\mathcal{E}$, a decoder $\mathcal{D}$, and a denoising U-Net $\mathcal{U}$ operating within the latent space. We begin by projecting an input image $x_0$ into the latent space through the encoder $\mathcal{E}$, resulting in a latent code $z_0$. Subsequently, we add Gaussian noise $\epsilon$ to the latent code according to a predefined time step $t$. Lastly, with the latent code $z_t$ at time step $t$, we then extract the SD features $\mathcal{F}_{SD}$ utilizing the denoising U-Net:
\begin{align}
    \mathcal{F}_{SD} = \mathcal{U}(z_t,t),\ z_t = \sqrt{\Bar{a}_t} + \sqrt{1-\Bar{a}_t}\epsilon,\ z_0=\mathcal{E}(x_0).
\end{align}

In accordance with the approach introduced in \cite{DBLP:journals/corr/abs-2303-02153}, we combine features extracted from different decoder $\mathcal{D}$ layers, specifically layers 2, 5, and 8, to capture multi-scale features. However, a direct concatenation of features from these three layers results in an excessively high-dimensional feature vector, approximately 5440 dimensions. To address this issue, we employ Principal Component Analysis (PCA) for each feature layer. Subsequently, we upsample lower-resolution features (i.e., layers 2 and 5) to match the resolution of the higher-resolution layer (i.e., layer 8) before concatenation.

\noindent \textbf{Feature Fusion.}
Expanding on the aforementioned discussions, we introduce a simple yet remarkably effective fusion strategy aimed at harnessing the advantages of both SD and DINO features. The core idea entails independently normalizing both sets of features to ensure consistency in their scales and distributions, subsequently concatenating them:
\begin{align}
    \mathcal{F}_{FUSE} =  Concat(\|\mathcal{F}_{SD}\|_2, \| 
 \mathcal{F}_{DINO}\|_2). 
\end{align}
In the third column of Fig. \ref{dinovis}, the fused feature aids in generating a smoother and more resilient group feature.

\subsection{Group Center Proxy Generation and TopK Selecting}
We acquire the feature $\mathcal{F}_{FUSE}  \in \mathbb{R}^{C \times H \times W}$ for each image, a valuable asset for robust group information generation. However, the subsequent challenge lies in disseminating the group information to individual images. In existing CoSOD methods, group information is often expressed as a feature map, directly concatenated with the original image. This is followed by a trainable decoder for co-saliency map prediction. In our proposed zero-shot framework, training a new decoder network using the CoSOD training dataset is not feasible. To tackle this challenge, we transform the representation of group information and introduce the pixel group center proxy generation and TopK selecting mechanism. Through this innovative approach, we can create group prompt points to represent co-salient objects in each images, which, in turn, prompt SAM to generate the corresponding maps.

Assuming an image group comprises $N$ images, we concatenate these features to generate the group feature $\mathcal{F}_{G} \in \mathbb{R}^{N\times C \times H \times W}$.  Subsequently, we reshape $\mathcal{F}_{G}$ into the shape $\mathbb{R}^{NHW \times C}$, denoting each pixel embedding as $\mathcal{F}_G^{ln}$, where $l \in [1,NHW]$ and $n$ means the $n$-th image. $\mathcal{F}_G^{ln}$ means that the $l$-th pixel embedding belongs to $n$-th image. Then, we use the easy averaging operation on these pixel embeddings to generate the group center proxy $\mathcal{F}_{c}$. Moreover, to make the group center proxy focus on salient regions, we use the unsupervised SOD method TSDN \cite{DBLP:conf/cvpr/Zhou0YLX23} to filter pixels belong to non-salient regions in $\mathcal{F}_G^{ln}$, generating the salient pixels $\mathcal{F}_G^{ln-s}$. The process of the generation of $\mathcal{F}_{c}$ is written as:
\begin{align}
     \mathcal{F}_{c} = Avg\left\{ \mathcal{F}_G^{ln-s}  \right\} \in \mathbb{R}^C. 
\end{align}
Concretely, for $N$-th image which contains $L$ salient pixels, we calculate the correlation score between $\mathcal{F}_{c}$ and $\mathcal{F}_G^{LN}$, and use TopK to select the point at position $P^N$ in the image $N$, which can represent common co-salient objects in this image: 
\begin{align}
     S^{LN} = \mathcal{F}_{c} \otimes \mathcal{F}^{LN}, P^N = {\text{TopK}}(S^{LN}) \in \mathbb{R}^K,
 \end{align}
where $\otimes$ means matrix multiplication. Finally, for the $N$-th image, the generated prompts at position $P^N$ (Fig. \ref{totalresults} and Fig. \ref{Vis}) and the corresponding original image is sent to CMP to generate the co-saliency maps. We set $K=2$ in this paper.

\section{EXPERIMENTS}
\subsection{Experimental Setup}

\noindent \textbf{Implementation Details.} 
We employ the SDv1-5 and DINOv2 models as our feature extractors, with the DDIM timestep $t = 50$ by default. We use the SAM with vit-b backbone. $N$ is the total image numbers of one certain image group. All experiments are conducted on a single  3090 GPU.

\begin{figure*}[!htbp]
    \centering
    \includegraphics[width=\linewidth]{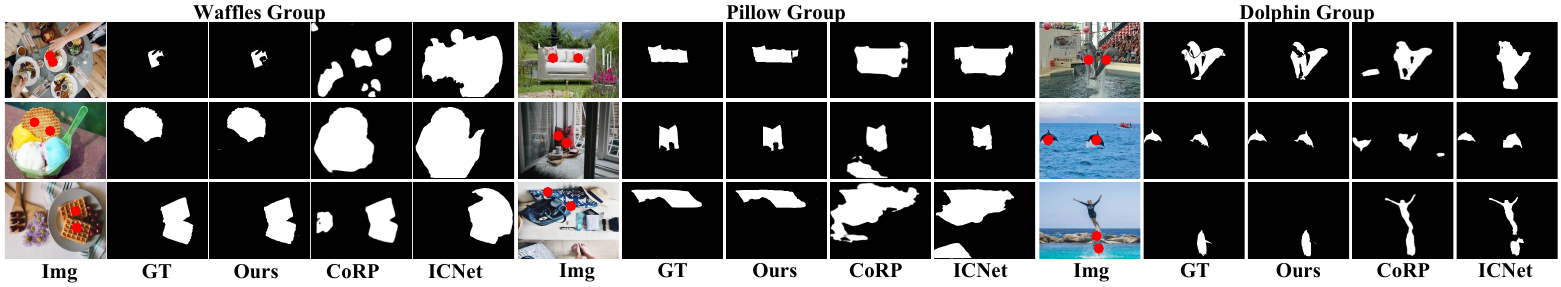}
    \caption{Visual comparison between our method and other methods. }
    \label{Vis}
    \vspace{-0.5cm}
\end{figure*}

\begin{table*}[!t]
\centering
\caption{Quantitative comparison with SOTA on  CoSOD datasets. The {\color[HTML]{FF0000} red} font means that our proposed network achieves the competitive performance compared to these methods.}
\scalebox{0.6}{
\begin{tabular}{c|c|c|c|ccc|ccc|ccc}
\hline
                                             & \cellcolor[HTML]{FFFFFF}                                   & \cellcolor[HTML]{FFFFFF}                                &                             & \multicolumn{3}{c|}{\cellcolor[HTML]{FFFFFF}\textbf{CoCA}}                                                                                                         & \multicolumn{3}{c|}{\cellcolor[HTML]{FFFFFF}\textbf{CoSOD3k}}                                                                                                      & \multicolumn{3}{c}{\cellcolor[HTML]{FFFFFF}\textbf{CoSal2015}}                                                                                                     \\ \cline{5-13} 
\multirow{-2}{*}{\textbf{Performance Level}} & \multirow{-2}{*}{\cellcolor[HTML]{FFFFFF}\textbf{Methods}} & \multirow{-2}{*}{\cellcolor[HTML]{FFFFFF}\textbf{Year}} & \multirow{-2}{*}{Type}      & \cellcolor[HTML]{FFFFFF}$S_m \uparrow$               & \cellcolor[HTML]{FFFFFF}$F^{mean}_\beta \uparrow$    & \cellcolor[HTML]{FFFFFF}$MAE \downarrow$             & \cellcolor[HTML]{FFFFFF}$S_m \uparrow$               & \cellcolor[HTML]{FFFFFF}$F^{mean}_\beta \uparrow$    & \cellcolor[HTML]{FFFFFF}$MAE \downarrow$             & \cellcolor[HTML]{FFFFFF}$S_m \uparrow$               & \cellcolor[HTML]{FFFFFF}$F^{mean}_\beta \uparrow$    & \cellcolor[HTML]{FFFFFF}$MAE \downarrow$             \\ \hline
                                             & TCNet                                                      & TCSVT 2023                                              & Supervised                  & {\color[HTML]{333333} 0.685}                         & {\color[HTML]{FE0000} 0.548}                         & {\color[HTML]{333333} 0.101}                         & {\color[HTML]{333333} 0.832}                         & {\color[HTML]{333333} 0.797}                         & {\color[HTML]{333333} 0.068}                         & {\color[HTML]{333333} 0.870}                         & {\color[HTML]{333333} 0.859}                         & {\color[HTML]{333333} 0.054}                         \\
                                             & \cellcolor[HTML]{FFFFFF}UFO                                & \cellcolor[HTML]{FFFFFF}TMM 2023                        & Supervised                  & \cellcolor[HTML]{FFFFFF}{\color[HTML]{333333} 0.697} & \cellcolor[HTML]{FFFFFF}{\color[HTML]{333333} 0.555} & \cellcolor[HTML]{FFFFFF}{\color[HTML]{333333} 0.095} & \cellcolor[HTML]{FFFFFF}{\color[HTML]{333333} 0.819} & \cellcolor[HTML]{FFFFFF}{\color[HTML]{333333} 0.783} & \cellcolor[HTML]{FFFFFF}{\color[HTML]{333333} 0.073} & \cellcolor[HTML]{FFFFFF}{\color[HTML]{333333} 0.860} & \cellcolor[HTML]{FFFFFF}{\color[HTML]{333333} 0.848} & \cellcolor[HTML]{FFFFFF}{\color[HTML]{333333} 0.064} \\
\multirow{-3}{*}{\textbf{Future}}            & \cellcolor[HTML]{FFFFFF}CoRP                               & \cellcolor[HTML]{FFFFFF}TPAMI 2023                      & Supervised                  & \cellcolor[HTML]{FFFFFF}{\color[HTML]{333333} 0.732} & \cellcolor[HTML]{FFFFFF}{\color[HTML]{333333} 0.603} & \cellcolor[HTML]{FFFFFF}{\color[HTML]{333333} 0.093} & \cellcolor[HTML]{FFFFFF}{\color[HTML]{333333} 0.850} & \cellcolor[HTML]{FFFFFF}{\color[HTML]{333333} 0.824} & \cellcolor[HTML]{FFFFFF}{\color[HTML]{333333} 0.057} & \cellcolor[HTML]{FFFFFF}{\color[HTML]{333333} 0.884} & \cellcolor[HTML]{FFFFFF}{\color[HTML]{333333} 0.884} & \cellcolor[HTML]{FFFFFF}{\color[HTML]{333333} 0.044} \\ \hline
                                             & \cellcolor[HTML]{FFFFFF}ICNet                              & \cellcolor[HTML]{FFFFFF}NeurIPS 2020                    & Supervised                  & \cellcolor[HTML]{FFFFFF}{\color[HTML]{FE0000} 0.651} & \cellcolor[HTML]{FFFFFF}{\color[HTML]{FE0000} 0.503} & \cellcolor[HTML]{FFFFFF}{\color[HTML]{FE0000} 0.148} & \cellcolor[HTML]{FFFFFF}{\color[HTML]{333333} 0.780} & \cellcolor[HTML]{FFFFFF}{\color[HTML]{333333} 0.734} & \cellcolor[HTML]{FFFFFF}{\color[HTML]{333333} 0.097} & \cellcolor[HTML]{FFFFFF}{\color[HTML]{333333} 0.856} & \cellcolor[HTML]{FFFFFF}{\color[HTML]{333333} 0.846} & \cellcolor[HTML]{FFFFFF}{\color[HTML]{333333} 0.058} \\
                                             & \cellcolor[HTML]{FFFFFF}CADC                               & \cellcolor[HTML]{FFFFFF}ICCV 2021                       & Supervised                  & \cellcolor[HTML]{FFFFFF}{\color[HTML]{333333} 0.681} & \cellcolor[HTML]{FFFFFF}{\color[HTML]{FE0000} 0.503} & \cellcolor[HTML]{FFFFFF}{\color[HTML]{FE0000} 0.132} & \cellcolor[HTML]{FFFFFF}{\color[HTML]{333333} 0.801} & \cellcolor[HTML]{FFFFFF}{\color[HTML]{333333} 0.742} & \cellcolor[HTML]{FFFFFF}{\color[HTML]{333333} 0.096} & \cellcolor[HTML]{FFFFFF}{\color[HTML]{333333} 0.866} & \cellcolor[HTML]{FFFFFF}{\color[HTML]{333333} 0.825} & \cellcolor[HTML]{FFFFFF}{\color[HTML]{333333} 0.064} \\
\multirow{-3}{*}{\textbf{Competitive}}       & \cellcolor[HTML]{FFFFFF}GLNet                              & \cellcolor[HTML]{FFFFFF}TCyb 2022                       & Supervised                  & \cellcolor[HTML]{FFFFFF}{\color[HTML]{FE0000} 0.591} & \cellcolor[HTML]{FFFFFF}{\color[HTML]{FE0000} 0.435} & \cellcolor[HTML]{FFFFFF}{\color[HTML]{FE0000} 0.188} & \cellcolor[HTML]{FFFFFF}{\color[HTML]{333333} -}     & \cellcolor[HTML]{FFFFFF}{\color[HTML]{333333} -}     & \cellcolor[HTML]{FFFFFF}{\color[HTML]{333333} -}     & \cellcolor[HTML]{FFFFFF}{\color[HTML]{333333} 0.855} & \cellcolor[HTML]{FFFFFF}{\color[HTML]{333333} 0.849} & \cellcolor[HTML]{FFFFFF}{\color[HTML]{333333} 0.060} \\ \hline
                                             & \cellcolor[HTML]{FFFFFF}CSMG                               & \cellcolor[HTML]{FFFFFF}CVPR 2019                       & Supervised                  & \cellcolor[HTML]{FFFFFF}{\color[HTML]{333333} 0.632} & \cellcolor[HTML]{FFFFFF}{\color[HTML]{333333} 0.494} & \cellcolor[HTML]{FFFFFF}{\color[HTML]{333333} 0.124} & \cellcolor[HTML]{FFFFFF}{\color[HTML]{333333} 0.711} & \cellcolor[HTML]{FFFFFF}{\color[HTML]{333333} 0.682} & \cellcolor[HTML]{FFFFFF}{\color[HTML]{333333} 0.157} & \cellcolor[HTML]{FFFFFF}{\color[HTML]{333333} 0.774} & \cellcolor[HTML]{FFFFFF}{\color[HTML]{333333} 0.775} & \cellcolor[HTML]{FFFFFF}{\color[HTML]{333333} 0.130} \\
                                             & PJO                                                        & TIP2018                                                 & Unsupervised                & {\color[HTML]{333333} 0.573}                         & {\color[HTML]{333333} 0.362}                         & {\color[HTML]{333333} 0.175}                         & {\color[HTML]{333333} 0.677}                         & {\color[HTML]{333333} 0.631}                         & {\color[HTML]{333333} 0.188}                         & {\color[HTML]{333333} 0.721}                         & {\color[HTML]{333333} 0.687}                         & {\color[HTML]{333333} 0.192}                         \\
                                             & GOMAG                                                      & TMM 2020                                                & Unsupervised                & {\color[HTML]{333333} 0.587}                         & {\color[HTML]{333333} 0.387}                         & {\color[HTML]{333333} 0.170}                         & {\color[HTML]{333333} 0.687}                         & {\color[HTML]{333333} 0.642}                         & {\color[HTML]{333333} 0.180}                         & {\color[HTML]{333333} 0.734}                         & {\color[HTML]{333333} 0.698}                         & {\color[HTML]{333333} 0.187}                         \\ \cline{2-13} 
                                             & Ours (-SD)                                                 &                                                         &                             & 0.653                                                & 0.532                                                & 0.121                                                & 0.717                                                & 0.679                                                & 0.127                                                & 0.776                                                & 0.787                                                & 0.110                                                \\
\multirow{-5}{*}{\textbf{Outperform}}        & \cellcolor[HTML]{FFFFFF}{\color[HTML]{000000} Ours}        & \multirow{-2}{*}{ICASSP 2023}                                  & \multirow{-2}{*}{Zero Shot} & \cellcolor[HTML]{FFFFFF}{\color[HTML]{000000} 0.667} & \cellcolor[HTML]{FFFFFF}{\color[HTML]{000000} 0.549} & \cellcolor[HTML]{FFFFFF}{\color[HTML]{000000} 0.115} & \cellcolor[HTML]{FFFFFF}{\color[HTML]{000000} 0.723} & \cellcolor[HTML]{FFFFFF}{\color[HTML]{000000} 0.691} & \cellcolor[HTML]{FFFFFF}{\color[HTML]{000000} 0.117} & \cellcolor[HTML]{FFFFFF}{\color[HTML]{000000} 0.785} & \cellcolor[HTML]{FFFFFF}{\color[HTML]{000000} 0.799} & {\color[HTML]{000000} 0.101}                         \\ \hline
\end{tabular}}
\label{totalresults}
\vspace{-0.5cm}
\end{table*}

\begin{table*}[!htbp]
\centering
\caption{Performance improvement of existing methods after adding group information extracted by DINO and SD. }
\scalebox{0.55}{
\begin{tabular}{@{}
>{\columncolor[HTML]{FFFFFF}}c 
>{\columncolor[HTML]{FFFFFF}}c |
>{\columncolor[HTML]{FFFFFF}}c 
>{\columncolor[HTML]{FFFFFF}}c 
>{\columncolor[HTML]{FFFFFF}}c |
>{\columncolor[HTML]{FFFFFF}}c 
>{\columncolor[HTML]{FFFFFF}}c 
>{\columncolor[HTML]{FFFFFF}}c |
>{\columncolor[HTML]{FFFFFF}}c 
>{\columncolor[HTML]{FFFFFF}}c c@{}}
\toprule
\multicolumn{1}{c|}{\cellcolor[HTML]{FFFFFF}}                                   & \cellcolor[HTML]{FFFFFF}                                       & \multicolumn{3}{c|}{\cellcolor[HTML]{FFFFFF}\textbf{CoCA}}                                                         & \multicolumn{3}{c|}{\cellcolor[HTML]{FFFFFF}\textbf{CoSOD3k}}                                    & \multicolumn{3}{c}{\cellcolor[HTML]{FFFFFF}\textbf{CoSal2015}}                                                   \\ \cmidrule(l){3-11} 
\multicolumn{1}{c|}{\multirow{-2}{*}{\cellcolor[HTML]{FFFFFF}\textbf{Methods}}} & \multirow{-2}{*}{\cellcolor[HTML]{FFFFFF}\textbf{Pub. \&Year}} & $S_m \uparrow$                       & $F^{mean}_\beta \uparrow$            & $MAE \downarrow$                     & $S_m \uparrow$                & $F^{mean}_\beta \uparrow$      & $MAE \downarrow$                & $S_m \uparrow$                 & $F^{mean}_\beta \uparrow$            & \cellcolor[HTML]{FFFFFF}$MAE \downarrow$ \\ \midrule
\multicolumn{1}{c|}{\cellcolor[HTML]{FFFFFF}TSCoSOD}                            & TIP 2022                                                       & 0.732 \textbf{(0.741)}                        & 0.613 \textbf{(0.625)}                        & 0.099 \textbf{(0.093)}                        & 0.843 \textbf{(0.856)}                 & 0.816 \textbf{(0.831)}                  & 0.062 \textbf{(0.052)}                   & 0.898 \textbf{(0.905)}                  & 0.897 (\textbf{0.909)}                        & \cellcolor[HTML]{FFFFFF}0.045 \textbf{(0.039)}    \\
\multicolumn{1}{c|}{\cellcolor[HTML]{FFFFFF}TCNet}                              & TCSVT 2023                                                     & 0.685 \textbf{(0.697)}                        & 0.548 \textbf{(0.561)}                        & 0.101 \textbf{(0.092)}                        & 0.832 \textbf{(0.843)}                 & 0.797 \textbf{(0.814)}                  & 0.068 \textbf{(0.061)}                   & 0.870 \textbf{(0.877)}                  & 0.859 \textbf{(0.877)}                        & \cellcolor[HTML]{FFFFFF}0.054 \textbf{(0.046)}    \\
\multicolumn{1}{c|}{\cellcolor[HTML]{FFFFFF}GCoNet+}                            & TPAMI 2023                                                     & {\color[HTML]{333333} 0.738 \textbf{(0.749)}} & {\color[HTML]{333333} 0.612 \textbf{(0.624)}} & {\color[HTML]{333333} 0.081 \textbf{(0.077)}} & 0.843 \textbf{(0.855)}                 & 0.813 \textbf{(0.827)}                  & 0.062 \textbf{(0.054)}                   & 0.881 \textbf{(0.891)}                  & {\color[HTML]{333333} 0.870 \textbf{(0.884)}} & \cellcolor[HTML]{FFFFFF}0.056 \textbf{(0.045)}    \\ \midrule
\multicolumn{2}{c|}{\cellcolor[HTML]{FFFFFF}Average Improvement}                                                                                 & {\color[HTML]{FE0000} +1.46\%}       & {\color[HTML]{FE0000} +2.05\%}       & {\color[HTML]{FE0000} +6.63\%}       & {\color[HTML]{FE0000} 1.41\%} & {\color[HTML]{FE0000} +1.86\%} & {\color[HTML]{FE0000} +13.11\%} & {\color[HTML]{FE0000} +2.69\%} & {\color[HTML]{FE0000} +1.63\%}       & {\color[HTML]{FE0000} +15.93\%}          \\ \bottomrule
\end{tabular}

}
\label{abla}
\vspace{-0.5cm}
\end{table*}

\noindent \textbf{Datasets.} We employ three benchmark datasets, including Cosal2015 \cite{DBLP:conf/cvpr/ZhangHLW15}, CoSOD3k \cite{9358006} and CoCA \cite{DBLP:conf/eccv/ZhangJXC20}, to evaluate our approach. Cosal2015 comprises 50 groups with a total of 2015 images. It presents numerous challenges such as complex environments. CoSOD3k, the largest-scale and most comprehensive benchmark, offers 160 groups with a total of 3000 images. CoCA features 80 groups with a total of 1297 images, posing a challenge due to the presence of multiple objects, including relatively small co-salient objects.

\noindent \textbf{Evaluation Metrics.}
we employ three widely used criteria: (1) F-measure $(F_\beta^{mean})$, representing the harmonic mean of precision and recall values, is calculated using a self-adaptive threshold. (2) Structure Measure $(S_m)$, which is utilized to assess the spatial structural similarities of saliency maps. (3) Mean Absolute Error $(MAE)$, which quantifies the average L1 distance between ground truth maps and predictions.

\subsection{Comparison Methods}
We compare our method with 7 fully-supervised methods: CSMG~\cite{DBLP:conf/cvpr/ZhangLL019}, ICNet~\cite{DBLP:conf/nips/Jin0CZG20}, CADC~\cite{DBLP:conf/iccv/ZhangHL021}, GLNet~\cite{GLNet}, TCNet \cite{DBLP:journals/tcsv/GeZXZB23}, CoRP~\cite{10008072} and UFO~\cite{su2023unified}. The unsupervised CoSOD models in barely explored, we only compare our method with 2 methods: PJO~\cite{DBLP:journals/tip/TsaiLHQL19} and GOMAG~\cite{DBLP:journals/tmm/JiangJTL21}.

\subsection{Quantitative and Qualitative Evaluation.} 
Table. \ref{totalresults} reveals that our proposed zero-shot CoSOD network consistently outperforms all other state-of-the-art (SOTA) unsupervised CoSOD methods across all evaluation metrics. These results underscore the efficacy of our zero-shot CoSOD network. In comparison to supervised CoSOD methods, our approach surpasses methods published in 2019 and achieves competitive performance compared to those published from 2020 to 2023, as evidenced by certain metrics. It's worth noting that all modules in our network employ basic design principles, such as simple averaging operations for center proxy point feature extraction. In this configuration, our zero-shot framework achieves such remarkable performance, instilling confidence in researchers to explore CoSOD tasks from a novel zero-shot perspective. Fig. \ref{Vis} presents qualitative results from our proposed method, showcasing its ability to accurately detect co-salient objects even in complex scenes.

\subsection{Ablation Analysis}
First, we propose that both high-level and low-level information are pivotal for generating group features. To validate this assertion, we conduct experiments. In the penultimate row of Table. \ref{totalresults}, utilizing only the high-level features extracted from DINO already achieves competitive performance within the proposed network. However, it does not enable our zero-shot framework to completely surpass  the CSMG method in all metrics. Additionally, the incorporation of low-level information from SD further enhances performance.

Another noteworthy contribution of this paper is the assertion that features extracted from foundational models are valuable for generating group features. Consequently, when we incorporate these group features into existing frameworks,
including TSCoSOD \cite{DBLP:journals/tip/0115TKSD22}, TCNet \cite{DBLP:journals/tcsv/GeZXZB23} and GCoNet+ \cite{zheng2022gconet+}, we observe further performance improvements, as demonstrated in Table. \ref{abla}. The values in parentheses represent results after retraining with the inclusion of new group features. This underscores that, beyond the framework itself, this paper contributes a zero-shot group feature generation approach.

\section{conclusion}

In this paper, we introduce an innovative zero-shot CoSOD framework. Leveraging the feature extraction capabilities of established DINO and SD, we have devised the GPG and CMP components, enabling the application of existing foundational models to the zero-shot CoSOD task. Our experiments demonstrate that these foundational models can effectively generate resilient group features, and our proposed framework can reasonably address the zero-shot CoSOD task. We envision that our work will serve as a cornerstone for the zero-shot/unsupervised CoSOD task, inspiring researchers to approach CoSOD from a novel perspective.

\clearpage
\bibliographystyle{IEEEbib}
\bibliography{strings,refs}
\end{document}